\documentclass[sigconf]{acmart}
\usepackage{multirow}

\renewcommand\footnotetextcopyrightpermission[1]{} 
\AtBeginDocument{%
  \providecommand\BibTeX{{%
    \normalfont B\kern-0.5em{\scshape i\kern-0.25em b}\kern-0.8em\TeX}}}

\newcommand{\zheng}{}

\begin{document}

\title{A Survey of Diffusion Based Image Generation Models: \\Issues and Their Solutions}

\author{Tianyi Zhang}
\affiliation{%
  \institution{Huawei Singapore Research Center}
  \country{Singapore}}
\email{zhang.tianyi@huawei.com}

\author{Zheng Wang}
\affiliation{%
  \institution{Huawei Singapore Research Center}
  \country{Singapore}}
\email{wangzheng155@huawei.com}

\author{Jing Huang}
\affiliation{%
  \institution{Huawei Singapore Research Center}
  \country{Singapore}}
\email{huangjing114@huawei.com}

\author{Mohiuddin Muhammad Tasnim}
\affiliation{%
  \institution{Huawei Singapore Research Center}
  \country{Singapore}}
\email{mohiuddin.muhammad.tasnim@huawei.com}

\author{Wei Shi}
\affiliation{%
  \institution{Huawei Singapore Research Center}
  \country{Singapore}}
\email{w.shi@huawei.com}

\renewcommand{\shortauthors}{Tianyi Zhang et al.}


\begin{abstract}
    Recently, there has been significant progress in the development of large models. Following the success of ChatGPT, numerous language models have been introduced, demonstrating remarkable performance. Similar advancements have also been observed in image generation models, such as Google's Imagen model, OpenAI's DALL$\cdot$E 2, and stable diffusion models, which have exhibited impressive capabilities in generating images. However, similar to large language models, these models still encounter unresolved challenges. Fortunately, the availability of open-source stable diffusion models and their underlying mathematical principles has enabled the academic community to extensively analyze the performance of current image generation models and make improvements based on this stable diffusion framework. This survey aims to examine the existing issues and the current solutions pertaining to image generation models.
\end{abstract}






\maketitle

\begin{table*}[h]

\centering
\caption{Summary of existing diffusion models in image generation tasks.}
\label{tab:summary}
\begin{tabular}{l|l|l}
\hline
Challenges                                               & Techniques                              & Models \\ \hline
\multirow{2}{*}{Multi-object generation}                 & Adding more controls                    & \begin{tabular}[l]{@{}l@{}}ReCo~\cite{Yang_2023_RECO}, GLIGEN~\cite{Li_2023_GLIGEN}, SceneComposer~\cite{zeng2022scenecomposer}, SpaText~\cite{Avrahami_2023_SpaText},\\  ControlNet~\cite{zhang2023adding}, UniControl~\cite{qin2023unicontrol}, Uni-ControlNet~\cite{zhao2023uni-control},\\   T2I-adapter~\cite{mou2023t2iadapter}, Universal Guidance~\cite{Bansal_2023_uguidance}, VPGen~\cite{Cho2023VPT2I} \\ \end{tabular}

        \\ \cline{2-3} 
                                                         & Working on attention maps               & \begin{tabular}[l]{@{}l@{}}SynGen~\cite{rassin2023linguistic}, Attend-and-Excite~\cite{chefer2023attendandexcite},\\ CTIS~\cite{wang2023compositional}, Detector Guidance~\cite{liu2023detector},\\ TFLA~\cite{mao2023trainingfree}, Attention-Refocusing~\cite{phung2023grounded}\end{tabular}
                                                         
                                                                 \\ \hline
\multirow{2}{*}{\begin{tabular}[l]{@{}l@{}}Generating rare \\or novel concepts\end{tabular}  } & Retrieval-based methods                 & RDM~\cite{blattmann2022semiparametric}, kNN-Diffusion~\cite{sheynin2023knndiffusion}, Re-imagen~\cite{chen2022reimagen}       \\ \cline{2-3} 
                                                         & Subject-driven generation               & \begin{tabular}[l]{@{}l@{}}Textrual Inversion~\cite{gal2022textual}, ELITE~\cite{wei2023elite}, Custom Diffusion~\cite{kumari2022customdiffusion},\\ DreamBooth~\cite{ruiz2022dreambooth}, E4T~\cite{gal2023encoderbased}, InstantBooth~\cite{shi2023instantbooth},\\ FastComposer~\cite{xiao2023fastcomposer}, Blip-Diffusion~\cite{li2023blipdiffusion}\end{tabular}
                                                         
                                                                  \\ \hline
\multirow{6}{*}{\begin{tabular}[l]{@{}l@{}}Quality improvement\\of generated images\end{tabular}}                     & Text encoder improvement                & Imagen~\cite{NEURIPS2022_imagen}, GlueGen~\cite{qin2023gluegen}, eDiff-i~\cite{balaji2022eDiff-I}, StructureDiffusion~\cite{feng2023structurediff}      \\ \cline{2-3} 
                                                         & Mixture of experts                      & eDiff-i~\cite{balaji2022eDiff-I}, ERNIE-ViLG~\cite{feng2023ernievilg}       \\ \cline{2-3} 
                                                         & Instruction tuning for human preference & HPS~\cite{wu2023better}, ImageReward~\cite{xu2023imagereward}       \\ \cline{2-3} 
                                                         & Sampling quality improvement            & Shifted Diffusion~\cite{zhou2022shifted}       \\ \cline{2-3} 
                                                         & Self-attention guidance                 & SAG~\cite{hong2022SAG}       \\ \cline{2-3} 
                                                         & Prompt rewriting                        & Promptist~\cite{hao2022Promptist}       \\ \hline
\end{tabular}
\end{table*}

\section{Introduction}
    {\zheng Text-to-image generation has been an ongoing challenge, with notable advancements made by Generative Adversarial Networks (GANs)~\cite{NIPS2014_GAN} and diffusion models~\cite{NEURIPS2020_DDPM}. Recently, OpenAI's DALL$\cdot$E 2~\cite{ramesh2022hierarchical} and Google's Imagen~\cite{chen2022reimagen} have showcased the impressive potential of AI in generating photorealistic images from text prompts. However, these models were not widely accessible as they were not open-sourced. Stability.AI's release of Stable Diffusion (SD) changed this landscape by offering an open-source diffusion model, where the open-source community creates user-friendly web interfaces and tutorials, making image generation accessible to individuals without expertise in the field. The diffusion-based image generation models became widespread, and fine-tuned models were shared on platforms like ``Civitai''.

    Despite the remarkable achievements of diffusion models, certain limitations persist in generating satisfactory results, including (1) the difficulty in generating images with multiple objects accurately, (2) challenges in generating rare or novel concepts, and (3) the need for consistent high-quality image generation. We provide more details for the challenges below, and summarize these works in Table~\ref{tab:summary}.}

    \begin{itemize}
        \item \textbf{C1: Difficulty generating images with multiple objects}: {\zheng Current generation models face challenges in accurately depicting images with multiple objects. An illustrated example (a yellow dog on the right and a black cat on the left) is shown in Figure~\ref{fig:multifail}. In this case, the generated images may completely omit the dog or incorrectly represent the color of the cat, leading to object omissions, attribute mismatches, or attribute mixing. On the other hand, current diffusion models struggle to capture positional information effectively.}
        \item \textbf{C2: Generating rare or novel concepts}: The performance of image generation models, similar to language models, heavily relies on the model and dataset size. {\zheng Despite having a large dataset, these models still struggle to generate rare or newly emerging concepts since they haven't encountered such examples during training.}
        
        \item \textbf{C3: Quality improvement of generated images}: Although {\zheng existing studies} often showcase impressive samples, these are often cherry-picked from numerous generated images or achieved after extensive prompt modifications. Generating {\zheng realistic and detailed images} that are consistently and efficiently within a few attempts remains a challenging task. 
    \end{itemize}
    \smallskip
    {\zheng \textbf{Overview.} In this survey, We organize the existing diffusion models for image generation by following the aforementioned challenges and their solutions. 
    Diffusion models are a class of generative models that adopt a unique approach to image generation by gradually adding noise to an image until it becomes completely degraded. The reverse process of removing noise aligns with the concept of image generation. Mathematically, this process is modeled as a Markov process (Section~\ref{sec:preliminaries}).
    For \textbf{C1}: To generate images with multiple objects, layout information such as bounding boxes or segmentation maps is added to the model. Cross-attention maps have been found to play a crucial role in image generation quality, and techniques like ``SynGen''~\cite{rassin2023linguistic} and ``Attend-and-Excite''~\cite{chefer2023attendandexcite} have been introduced to improve attention maps (Section~\ref{sec:c1}). 
    %
    For \textbf{C2}: The main approach is to incorporate with retrieval system. This method proposed to get rare or unseen abject information from retrieval system to introduce the new concepts to the model. Given the images of the objects, subject-driven generation is a task to generate images according to these given images. Various techniques, including fine-tuning and leveraging image encoders, are explored to enhance the model's capability in producing high-quality images without extensive test-time training (Section~\ref{sec:c2}). 
    For \textbf{C3}: Improvements in diffusion models have been achieved through various means. Enhancing the text encoder has been shown to be crucial for text-to-image generation, and works like ``Imagen'' have studied the impact of different text encoders. ``Mixture of experts'' has been used to leverage different models for different stages of the generation process. ``Instruction tuning for human preference'' employs reinforcement learning with a reward model to optimize image quality based on human preferences. Additionally, ``Sampling quality improvement'' and ``Rewrite the prompt'' have been introduced as techniques to enhance image generation quality (Section~\ref{sec:c3}).
    In general, diffusion models have made significant progress in generating high-quality images with diverse conditions. Researchers continue to explore novel techniques and improvements to further advance the capabilities of diffusion models in image generation tasks.

    The survey is different with existing surveys in several aspects. (1) Application Focus: Unlike the other surveys~\cite{10081412, yang2023diffusion, cao2023survey} that provide an overview of diffusion models and their mathematical foundations, our survey focuses on the application of diffusion models in image generation. It concludes the methods and techniques used to enhance image generation quality, which is essential for many practical applications
    (2) Emphasis on Limitations and Solutions: While existing surveys~\cite{10081412, yang2023diffusion, cao2023survey, zhang2023texttoimage} mostly introduce diffusion models and their applications, this survey takes a more targeted approach by identifying the current limitations of diffusion models in image generation. It then highlights individual research papers and their contributions in addressing these limitations. This enables readers to understand how various works have tackled specific challenges related to image generation using diffusion models.
    (3) Complementary to Other Modalities: Unlike the existing surveys that focus on the applications of diffusion models in other modalities, such as audio~\cite{zhang2023audio}, graph~\cite{liu2023generativegraph}, and medical images~\cite{kazerouni2023medicalimagediffusion}, this survey complements them by providing a comprehensive exploration of image generation using diffusion models. It consolidates the advancements made in this field, offering insights for researchers and practitioners interested in image synthesis.    
    }
    
\section{Preliminaries of diffusion models}
\label{sec:preliminaries}
    Prior to the emergence of diffusion models~\cite{pmlr-v37-sohl-diffusion}, GAN~\cite{NIPS2014_GAN} based or VAE~\cite{VAE} based methods are all tried to reconstruct images from sampled white noise. Building upon this concept, diffusion models~\cite{pmlr-v37-sohl-diffusion} took inspiration from thermodynamics and proposed a different approach. They reasoned that by gradually adding noise to an image until it becomes completely degraded into noise, the reverse process of adding noise aligns with the fundamental concept of image generation. If a deep neural network model can effectively capture and model this reversed process, it becomes possible to iteratively remove the noise, eventually revealing a clear image. This forms the fundamental idea behind diffusion models.

    \begin{figure*}[h!]
        \centering
        \includegraphics[width=\textwidth]{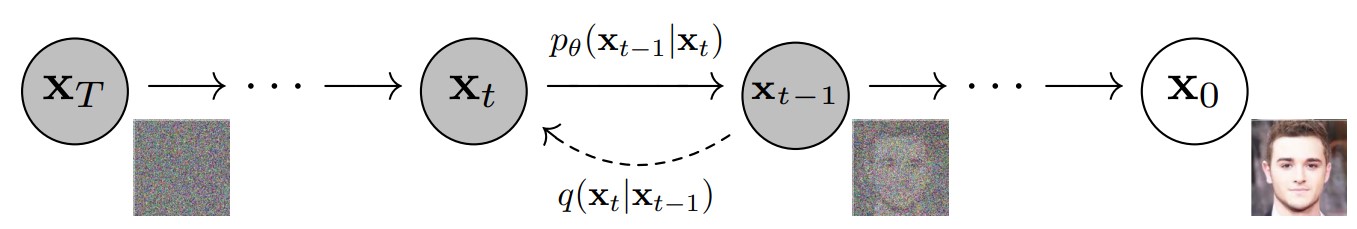}
        \caption{The directed graphical model considered in DDPM~\cite{NEURIPS2020_DDPM}.}
        \label{fig:ddpm}
    \end{figure*}

    Mathematically, this process can be modeled as a Markov process. Fig.\ref{fig:ddpm} is used in paper DDPM~\cite{NEURIPS2020_DDPM} to illustrate this process. $X_0$ represents the original image, $X_1$,$\dots$, $X_{t-1}$, $X_t$,$\dots$, $X_T$ represent the results after adding noise at each step. From fig \ref{fig:ddpm}, we can see that, at step $t-1$, $X_{t-1}$ is already a mixture of noise and image. After multiple rounds of injecting the noise, at step $T$, the image $X_T$ has already become something too noisy that can not be recognized anything. The process of adding noise step by step from $X_0$ to $X_T$ is called ``forward process'' or ``diffusion process''. In fig \ref{fig:ddpm}, $q(X_t|X_{t-1})$ refers to this process. Reversely, from $X_T$, the process of iterative remove the noise until getting the clear image is called the ``reverse process''. In fig \ref{fig:ddpm}, $p_\theta(X_t|X_{t-1})$ refers to this reverse process. 

    In the context of the forward process, the addition of noise is precisely described by equation \ref{forwardpass}. The variable $\epsilon_{t-1}$ represents the added noise. This noise is introduced through a weighted summation, where the weights at each step are denoted as $\beta_t$ or referred to as the diffusion rate. The determination of this diffusion rate is performed in advance through the utilization of a scheduler.
    
    \begin{equation}\label{forwardpass}
        x_t = \sqrt{1-\beta_t} \cdot x_{t-1} + \sqrt{\beta_t} \cdot \epsilon_{t-1}  
    \end{equation}

    Given this definition, mathematically, we can have equation \ref{target}, $\widetilde{\mu_t}(x_t,x_0)$ is in equation \ref{mu_tm1}. All the $alpha$ can be calculated using $\beta$ (detailed derivation can refer to~\cite{NEURIPS2020_DDPM}). The conclusion here is, given the original image $x_0$ and $x_t$, we can use equation \ref{target} to get $x_{t-1}$. Now if we have a neural network $\mu_\theta(x_t,t)$ to mimic this function, we can have our loss function in equation \ref{loss}. This is the original idea of diffusion model.
    
    \begin{equation}\label{target}
        q(X_{t-1}|X_t,X_0) =  \mathcal{N}(X_{t-1} ; \widetilde{\mu_t}(x_t,x_0),\widetilde{\beta_t} I)
    \end{equation}
    
    \begin{equation}\label{mu_tm1}
        \widetilde{\mu_t}(x_t,x_0) = \frac{\sqrt{\overline{\alpha}_{t-1}}}{1-\overline{\alpha}_t} \cdot x_0 + \frac{\sqrt{\alpha}_t\cdot(1-\overline{\alpha}_{t-1})}{1-\overline{\alpha}_t}\cdot x_t
    \end{equation}
    
    \begin{equation}
        \widetilde{\beta_t} = \frac{1-\overline{\alpha}_{t-1}}{1-\overline{\alpha}_t} \cdot \beta_t
    \end{equation}
    
    \begin{equation}\label{loss}
        L=\| \widetilde{\mu_t}(x_t,x_0)-\mu_{\theta}(x_t,t)\|^2
    \end{equation}

    Denoising Diffusion Probabilistic Models (DDPM)~\cite{NEURIPS2020_DDPM} further improved this idea. They proposed to directly predict the noise added at each step, $\epsilon_{t-1}$, $X_{t-1}$ can be easily obtained by removing $\epsilon_{t-1}$ from $X_t$. The proposed loss shows in equation \ref{loss_ddpm} DDPM~\cite{NEURIPS2020_DDPM} is the first diffusion model successfully generating high quality images. 
     
     \begin{equation}\label{loss_ddpm}
         L=\| \epsilon-\epsilon_{\theta}(x_t,t)\|^2
     \end{equation}

    DDPM originally focused on image generation from noise; however, real-world scenarios often require image generation conditioned on additional information. To address this need, Classifier-free guidance (CFG)~\cite{ho2022classifierfree} was proposed as a framework for effectively incorporating conditional inputs. This framework allows precise control over the conditioning information by changing the guidance rate. Since its introduction, the CFG framework has gained widespread adoption and serves as the foundational basis for numerous contemporary diffusion models, including Imagen~\cite{NEURIPS2022_imagen}, DALL$\cdot$E 2~\cite{ramesh2022hierarchical}, Stable Diffusion~\cite{Rombach_2022_LDM_CVPR}.
    
    Now we can summarize the basic components of a diffusion model:
    \begin{itemize}
        \item \textbf{Noise prediction module}: This is the core component of the diffusion model, taking $X_t$ or together with other condition as input, the output is the added noise. Due to the success of U-net~\cite{unet} in previous computer vision tasks, normally diffusion models also use this structure as their noise prediction model.~\cite{Rombach_2022_LDM_CVPR} also added attention layer~\cite{attention} to involve other conditions such as text or bounding boxes. Surprisingly, a very important feature is observed on this attention U-Net, the cross attention map in the U-net will directly affect the layout and geometry of the generated image~\cite{hertz2022prompt}. This feature is used by a lot of researchers to improve the generation quality. Recently there are also some other works replace this U-net with pure transformer structure~\cite{vitbackbone}. The main purpose of this component is to take some condition information and previous state as input, and output the next state with same dimension with previous one. 
        \item \textbf{Condition encoder}: Currently, most of the works are not just generate images from sampled noise, but also need conditioned on something, such as text. Text-to-image generation is the most common case. Same as all the other pretrained large multi-modal models, using some pretrained model and freezing it during the training is always a good strategy to save time. Current diffusion models also used this methodology. Normally, T5~\cite{2020t5} series encoder or CLIP~\cite{pmlr-v139-clip} text encoder is used in most of the current works.
        \item \textbf{Super resolution module}: most generation models typically operate on lower resolutions such as $512\times512$ or $256\times256$. However, real-world applications often require higher-resolution images. Take the example of wallpapers, which are commonly expected to be in 4K resolution. To address this limitation, a common approach is to incorporate super-resolution models after the initial generation process. For instance, DALL$\cdot$E 2~\cite{ramesh2022hierarchical} employs two super-resolution models in its pipeline. Initially, it generates an image of dimension $64\times64$. Subsequently, the first super-resolution model upscales the image to $256\times256$, and the second super-resolution model further enhances the image to $1024\times1024$. The inclusion of super-resolution modules not only increases the image size but also improves the image quality by adding more intricate details. As a result, features like facial pores and wrinkles may appear in the generated face after super-resolution, contributing to a more realistic image appearance. 
        \item \textbf{Dimension reduction module}: To save computational resources,  diffusion models often operate not directly on the pixel space of images, but rather on a lower-dimensional latent space representation. For instance, SD model~\cite{Rombach_2022_LDM_CVPR} utilizes VAE~\cite{VAE} to compress the image into a latent space of dimension $4\times64\times64$. On the other hand, DALL$\cdot$E 2~\cite{ramesh2022hierarchical} takes a different approach by performing the diffusion process directly on the embedding space of~\cite{pmlr-v139-clip}. In this case, both the text encoder and image encoder of CLIP~\cite{pmlr-v139-clip} are components integrated into the DALL$\cdot$E 2 model. 
        \item \textbf{Scheduler}: as mentioned previously, this module will define the diffusion rate at each step, $\beta_t$.
    \end{itemize}
    
\section{Multi-Objects generation}
\label{sec:c1}
\begin{figure}[t]
        \centering
        \includegraphics[width=0.4\textwidth]{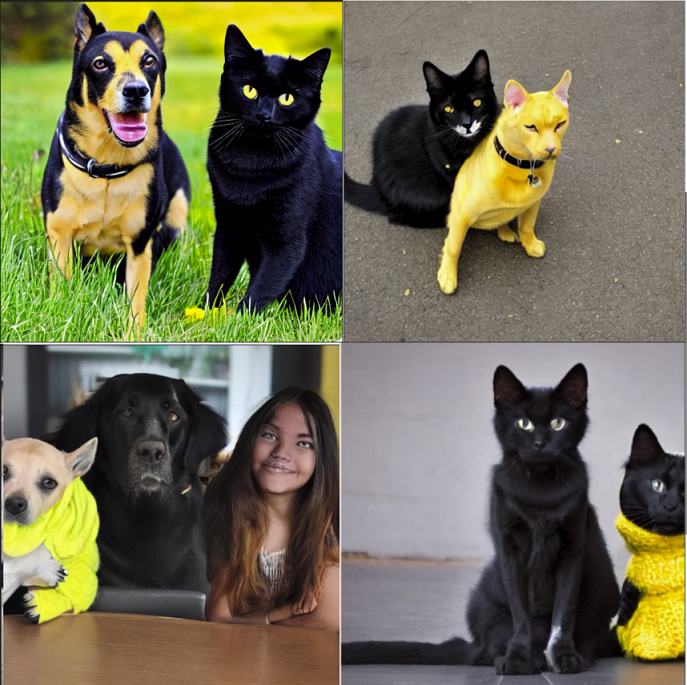}
        \caption{Failed cases for multiple objects generation.}
        \label{fig:multifail}
    \end{figure}

    When generating images with multiple objects, various issues often arise. In Figure \ref{fig:multifail}, we present four images generated by SD v1.5, illustrating the challenges encountered. In the upper two images, the prompt used was ``a yellow dog and a black cat.'' It is evident that both images have failed to accurately represent the intended objects. In the upper-left image, the dog exhibits some black coloration, while the upper-right image shows a yellow creature with mixed characteristics of both a dog and a cat. The bottom two images were generated with a more challenging prompt, ``a yellow dog, a black cat, and a girl.'' In both images, it is evident that the generated features do not align with the specified objects, and the bottom-right image completely lacks the presence of a girl. These examples exemplify common failure cases characterized by attributes mismatching, attributes mixing, and objects missing.
    
\begin{figure}[t]
        \centering
        \includegraphics[width=0.45\textwidth]{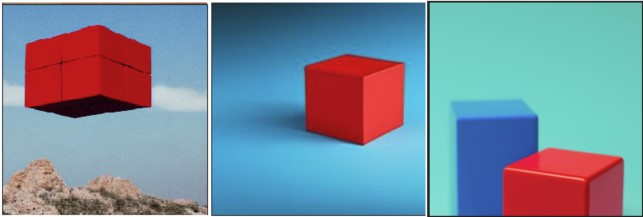}
        \caption{Failed cases for multiple objects generation: wrong position.}
        \label{fig:multifail2}
    \end{figure}

    Diffusion models can also encounter difficulties in accurately representing positional information. In fig \ref{fig:multifail2}, we present examples of this failure. The prompt used in all three images is ``a red cube on top of a blue cube.'' The left image is generated using SD v1.5, the middle image using the open-sourced unCLIP structure model Karlo~\cite{kakaobrain2022karlo-v1-alpha}, and the right image using DeepFloyd IF (IF), which is an open-sourced model based on Imagen~\cite{NEURIPS2022_imagen} and regarded as one of the best open-sourced text-to-image models available. In the first two images, the issue of objects missing persists, as the red cube is not present. In the IF-generated image, both the blue and red cubes are successfully generated. However, the positional information is incorrect, as the cubes are not in the intended arrangement.

    This showcases a common failure case in diffusion models, where accurately representing the correct positional information remains a challenge.
    
    Based on these findings, the following part of this section will introduce the proposed solutions.
    
\subsection{Adding More Control}
    One of the primary approaches to address this issue involves incorporating layout information into the model. This layout information can take the form of bounding boxes for individual objects or segmentation maps. In original Latent Diffusion Model (LDM) paper~\cite{Rombach_2022_LDM_CVPR}, the conditioning was not restricted to text, making it a versatile framework applicable to various domains.  Nevertheless, training such models from scratch can be computationally intensive. As a result, many recent works have adopted fine-tuning techniques to augment the model with additional conditioning based on pretrained models, often utilizing SD~\cite{Rombach_2022_LDM_CVPR} as a base. ReCo~\cite{Yang_2023_RECO} and GLIGEN~\cite{Li_2023_GLIGEN} both incorporating bounding box information into the model. ReCo~\cite{Yang_2023_RECO} involves extensive fine-tuning on SD 1.4, introducing extra position token embeddings into the original CLIP~\cite{pmlr-v139-clip} text encoder to encode the bounding box coordinates. Both the text encoder and the diffusion module are fine-tuned in this approach.  On the other hand, GLIGEN~\cite{Li_2023_GLIGEN} adopts common fine-tuning method from natural language processing (NLP), incorporating an adaptor layer between the self-attention and cross-attention layers.

    \begin{figure}[t]
        \centering
        \includegraphics[width=0.45\textwidth]{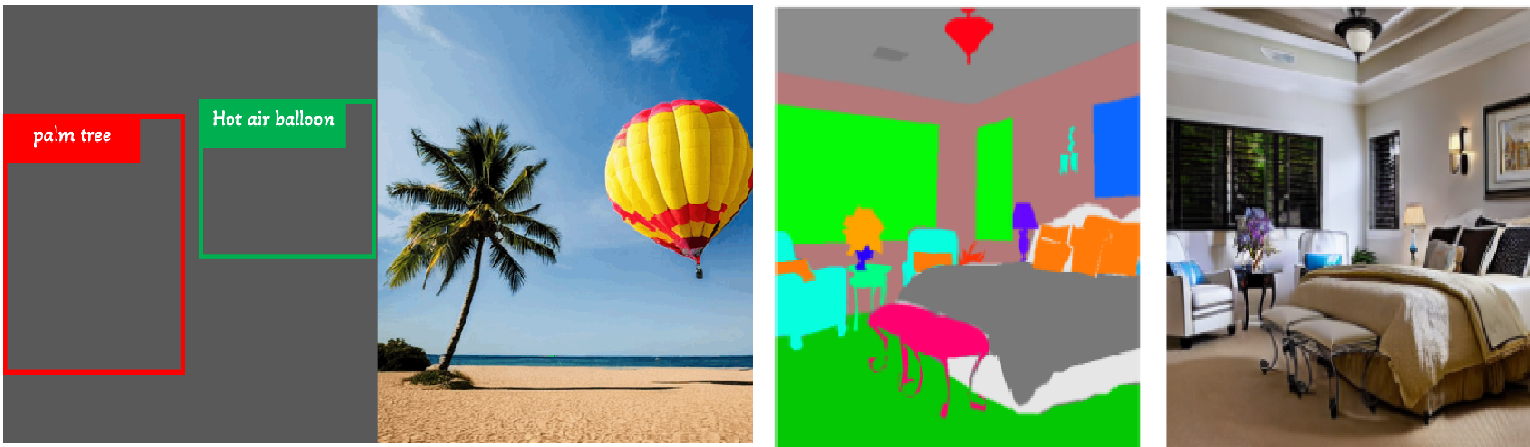}
        \caption{Bounding box to image and segmentation map to image~\cite{Li_2023_GLIGEN}.}
        \label{fig:bounding box and segmentation map to image}
    \end{figure}
    
    The utilization of segmentation maps has proven to be beneficial. Two notable works, SceneComposer~\cite{zeng2022scenecomposer} and SpaText~\cite{Avrahami_2023_SpaText}  concentrate on leveraging segmentation maps for image synthesis. SceneComposes~\cite{zeng2022scenecomposer} introduces an encoding method that accommodates imprecise segmentation maps, enabling image generation with more flexibility. On the other hand, SpaText~\cite{Avrahami_2023_SpaText} employs a specifically designed encoding mechanism to effectively capture and represent segmentation information.  ControlNet~\cite{zhang2023adding} stands as a prominent work in this domain. It introduces a parallel network alongside the unet~\cite{unet} architecture of SD~\cite{Rombach_2022_LDM_CVPR}, serving as a plug-in that imparts remarkable flexibility to the system. This structure has gained widespread adoption in the open-source community. Apart from segmentation maps, ControlNet~\cite{zhang2023adding} can also incorporate other types of input, such as depth maps, normal maps, canny maps, and MLSD straight lines. Several other works, including  Uni-ControlNet~\cite{zhao2023uni-control}, UniControl~\cite{qin2023unicontrol}, T2I-Adapter~\cite{mou2023t2iadapter} have followed suit by integrating various conditional inputs and incorporating additional layers to facilitate seamless connectivity between the conditioning information.

    Instead of the directly using the layout info as model input, Universal Guidance~\cite{Bansal_2023_uguidance} took the idea of classifier guidance~\cite{dhariwal2021diffusion}, extended the condition to all the other cases during the inference process. 

    The integration of layout information has demonstrated notable improvements in generating multiple objects in image synthesis, however, these are all based on the premise we already have the layout info. A significant challenge lies in generating the layout information itself. Several approaches have been explored to address this issue. One straightforward method involves employing large language models (LLMs) to generate bounding boxes~\cite{phung2023grounded,Cho2023VPT2I,lian2023llmgrounded}. This approach directly prompts the LLMs to generate the coordinates for the bounding boxes, albeit requiring careful engineering of the prompts. One such approach, VPGen~\cite{Cho2023VPT2I} took the advantage of LLMs. VPGen is short form for visual programming. It capitalizes on LLMs to accomplish this task. Rather than directly generating the image, VPGen employs an LLM to analyze the prompt. It first determines the number of objects to be included in the image, then proceeds to generate coordinates for each object. Finally, it feeds all the layout information to GLIGEN~\cite{Li_2023_GLIGEN} to generate the final image. Another method to generate layout information is by training a separate model dedicated to bounding box generation~\cite{liu2023detector,wang2023compositional}.

    Overall, methods mentioned in this section may require retraining specific parts or the entire model. This process can be costly and time-consuming. 
    
\subsection{Working on the Attention Maps}

    \begin{figure*}[h!]
        \centering
        \includegraphics[width=\textwidth]{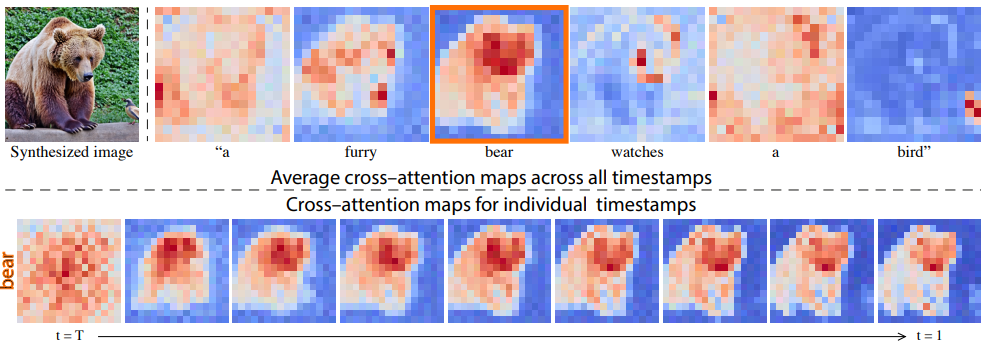}
        \caption{Cross-attention maps of a text-conditioned diffusion image generation~\cite{hertz2022prompt}.}
        \label{fig:cross-attention}
    \end{figure*}

\begin{figure}[h!]
        \centering
        \includegraphics[width=0.45\textwidth]{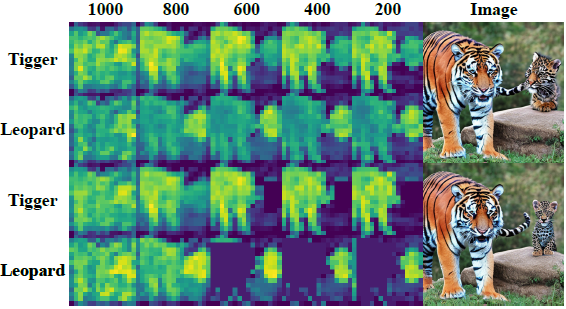}
        \caption{The Cross attention maps at different time steps, with and without correction. Prompt is 
 ``A striped tiger and a spotted leopard''~\cite{liu2023detector}.}
        \label{fig:CAMs}
    \end{figure}

    2022, Google research published ``Prompt-to-prompt''~\cite{hertz2022prompt}. In this paper, they mentioned one very important observation, which is the spatial layout and geometry of the generated image depend on the U-net's cross-attention maps. Fig \ref{fig:cross-attention} is the illustration of this observation.The top part of the figure demonstrates how the attention map corresponding to the ``bear'' token captures the location and shape of the final generated bear in the image. Moreover, the bottom part of the figure reveals that the structural characteristics of the image are already determined in the early stages of the diffusion process. Subsequent work, ``Attend-and-Excite''~\cite{chefer2023attendandexcite}, have further emphasized the importance of generating accurate cross-attention maps during the inference phase. These studies have established a crucial finding that if the model fails to generate appropriate cross-attention maps, the resulting generated image will likely be of poor quality. Fig \ref{fig:CAMs}, illustrates how erroneous attention maps can mislead the image generation process. From the attention maps, the top two rows of fig \ref{fig:CAMs}, ``tiger'' and ``leopard'' shared similar attention maps, resulting in an image with mixed attributes. The object on the right exhibits the head of a tiger but has the spotted fur texture associated with a leopard.
    
    Building upon this observation, researchers have proposed methods that directly manipulate the attention maps to generate higher-quality images. Figure \ref{fig:CAMs} illustrates this concept, where the corrected attention maps for ``tiger'' and ``leopard'' exhibit distinct hotspots. By correcting the attention maps, the resulting image aligns with the intended prompt.  

    ``SynGen'' model~\cite{rassin2023linguistic} works on cross attention map to overcome the attribute mixing issue. They stated that, the object and its corresponding attention map should share the same hot area. For example, for prompt ``a red apple and a yellow banana'', the attention maps of token ``red'' and ``apple'' should looks similar, so do ``yellow'' and ``banana''. This work proposed to use the similarity score of object and its attribute attention maps as optimization object, to force the object and its attribute to share the same hot area in the attention maps.

    The layout information can also help in the case. Several recent works~\cite{liu2023detector, mao2023trainingfree, chen2023trainingfree, phung2023grounded, wang2023compositional} have focused on enhancing cross attention maps by incorporating information from provided bounding boxes. Their primary objective is to augment attention values within specified bounding boxes while suppressing attention values outside of them. For example, as shown in Figure \ref{fig:CAMs}, when the prompt ``A striped tiger and a spotted leopard'' is considered, the attention map for the ``tiger'' token exhibits increased intensity within the ``tiger'' region, while the initial prominent area corresponding to the ``leopard'' region diminishes. These methodologies have exhibited significant improvements in the generation quality compared to the original SD model.

    Instead directly modify the attention map, Attend-and-Excite~\cite{chefer2023attendandexcite} proposed to iterative modify the latent to achieve better attention map. They called it ``On the Fly Optimization''. Intuitively, each subjects should have an area of high attention value on the corresponding attention map. The optimization objective encouraged this. The loss function is designed as:
    
    \begin{equation}
        \mathcal{L} = \mathop{max} \limits_{s\in S} \mathcal{L}_s \quad where \quad \mathcal{L}_s = 1-max(A^s_t)
    \end{equation}
    
    This loss function provided the direction to shift the latent $z_t$, the latent optimization function is $z_t^\prime \leftarrow z_t - \alpha_t \nabla_{z_t} \mathcal{L}$. After optimizing several iteration on this formula, the latent should be shifted to the one with better desired attention map, which is at least one hot area should appear in the object tokens' attention map. This $z_t^\prime$ will be used as input of the noise prediction U-net to predict $z_{t-1}$. So although optimization is called, there is no finetune on the model.

    Compared with methods mentioned in previous section, methods mentioned in this section is normally cheap because they are training free. 
    
\section{Rare object or unseen object generation}
\label{sec:c2}
    Similar to LLMs,image generation models also encounter challenges when generating rare or unseen objects. To address this issue, a common approach is to leverage search engines. In the context of LLMs, if the model is asked about the final score of a recently completed basketball game, it typically resorts to a search engine, retrieves the relevant information online, and subsequently paraphrases it to generate the final answer. Similarly, image generation models can derive advantages from retrieved information to enhance their generation capabilities.

\subsection{Retrieval Based Methods}

    The RDM model, as described in~\cite{blattmann2022semiparametric} capitalizes on the capabilities of CLIP~\cite{pmlr-v139-clip}.  In this approach, the condition model utilizes CLIP~\cite{pmlr-v139-clip} encoders. During the training process, CLIP~\cite{pmlr-v139-clip} is employed to retrieve neighboring images from a database. Subsequently, the CLIP embeddings of these retrieved images are used as condition inputs for the diffusion model. In the inference phase, the input is highly flexible. It is assumed that the text embedding should match the corresponding image embedding. Consequently, during inference, retrieval can be employed to obtain similar images, or the text embedding directly derived from the CLIP encoder can be used as input. 
    
    Another related work, kNN-Diffusion~\cite{sheynin2023knndiffusion} shares similarities with RDM. In this work, text is utilized to retrieve the top k nearest neighbor images from the database. These retrieved images, along with the text itself, are used as input during both the training and inference stages.

    \begin{figure*}[h!]
        \centering
        \includegraphics[width=\textwidth]{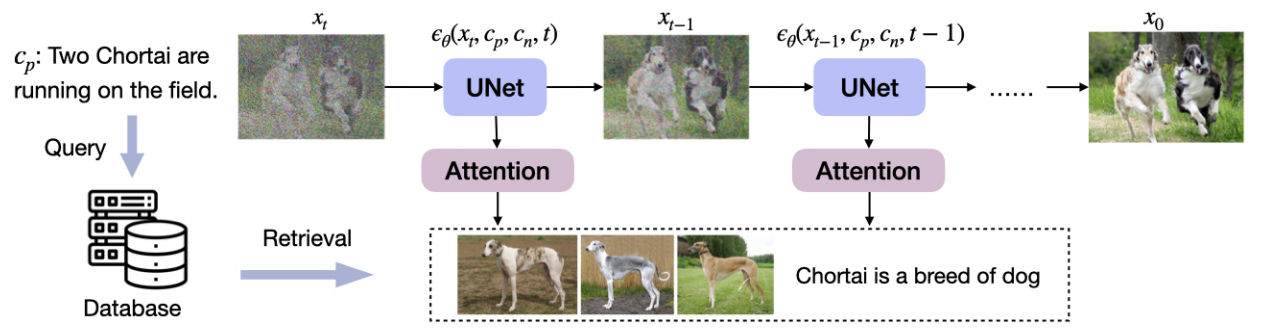}
        \caption{Pipeline of Re-imagen~\cite{chen2022reimagen}.}
        \label{fig:Reimagen}
        \vspace{-2mm}
    \end{figure*}

    Both RDM~\cite{blattmann2022semiparametric} and kNN-Diffusion~\cite{sheynin2023knndiffusion} utilize retrieved images as condition inputs during their respective processes. On the other hand, Re-imagen~\cite{chen2022reimagen} takes a different approach by conditioning its generation on both retrieved texts and images. For instance, as shown in fig \ref{fig:Reimagen}, when generating based on the prompt ``Two Chortai are running on the field,'' the retrieval of text is also taken into consideration. As a result, one of the top-n retrieved conditions could consist of both the text ``Chortai is a breed of dog'' and its corresponding image. This joint conditioning on both text and image retrieval allows Re-imagen to incorporate additional context and information, potentially leading to more contextually relevant and accurate generations.

\subsection{Subject Driven Generation}

\begin{figure}[h!]
        \centering
        \includegraphics[width=0.45\textwidth]{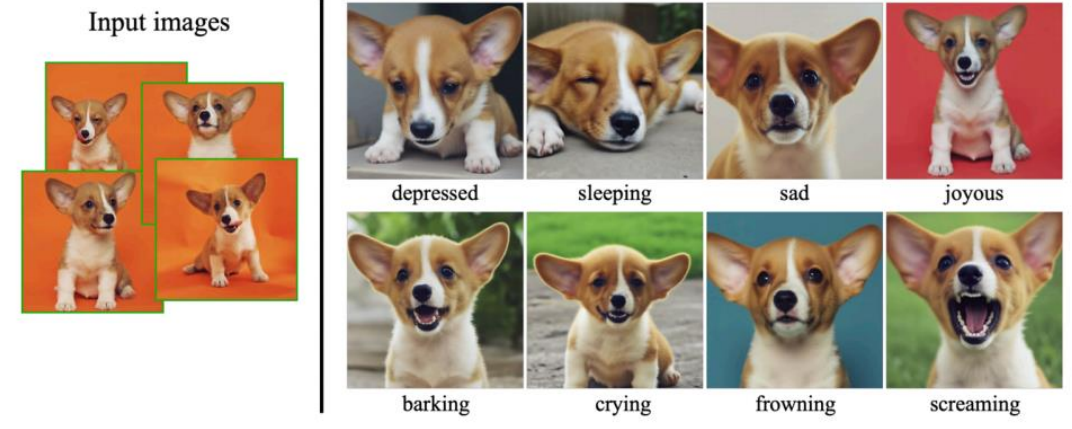}
        \caption{Subject driven image generation~\cite{ruiz2022dreambooth}.}
        \label{fig:subject-driven}
        \vspace*{-3mm}
    \end{figure}

    The retrieval based methods discussed earlier have demonstrated the utilization of retrieved images as inputs to introduce novel concepts. This approach can be generalized and extended to address another task known as subject-driven image generation, which is also referred to as concept customization or personalized generation. In subject-driven image generation, the main objective is to present an image or a set of images that represent a particular concept, and then generate new images based on that specific concept. Figure. \ref{fig:subject-driven}  illustrates this concept, where four images of a single dog are provided as input, and the generated images are all centered around the concept represented by that particular dog. 

    Subject-driven image generation allows for a more interactive and personalized image generation process, where users can input specific images that encapsulate the desired concept, leading to image generation that aligns closely with their intended subject or theme. This approach opens up exciting possibilities for customizing image generation outputs according to user-defined criteria and preferences.

    To incorporate subject-driven image generation into the existing framework, a feasible solution is to introduce a special token, often referred to as a unique identifier. This special token can be used in various ways, such as fine-tuning the embedding of the token on the given target concept images or part of the diffusion model. By doing so, the model can be trained to specialize or overfit on the provided concept, effectively embedding the concept's representation into this special identifier.
    
    Several methods have adopted this approach, including Textual Inversion~\cite{gal2022textual}, ELITE~\cite{wei2023elite}, Custom Diffusion~\cite{kumari2022customdiffusion}, DreamBooth~\cite{ruiz2022dreambooth}, E4T~\cite{gal2023encoderbased}. They leverage fine-tuning or specialized training techniques to embed the target concept within the special token, allowing the model to generate images closely aligned with the given subject or theme. Imagic~\cite{kawar2023imagic} approach utilizes a more intricate methodology but achieved better result. But still the model is finetuned. 

    Indeed, many of the methods discussed earlier require fine-tuning the model based on the given images, which can be time-consuming and computationally expensive. However, there are alternative approaches that aim to avoid test-time training and streamline the generation process. InstantBooth~\cite{shi2023instantbooth} and FastComposer~\cite{xiao2023fastcomposer} utilize a different image encoder to encode the concept information, which is then combined with the text embedding. Subsequently, the diffusion model is fine-tuned to accommodate this new condition, allowing for concept-specific image generation without the need for extensive test-time training. FastComposer~\cite{xiao2023fastcomposer} takes this concept-driven generation a step further by enabling image generation based on multiple concepts and also learning their spatial locations. Blip-diffusion~\cite{li2023blipdiffusion} is another work of blip-series~\cite{li2022blip,instructblip,li2023blip2}, which is well-known for its ``Q-former'' connector used to align text and images. Blip-diffusion also employs this Q-former to transfer concept images to the text domain. The advantage of Blip-diffusion lies in its ability to perform ``zero-shot'' generation, as well as ``few-shot'' generation with minimal fine-tuning.

    By incorporating these innovative approaches, subject-driven image generation can be achieved efficiently and effectively, enabling the generation of images based on specified concepts without the need for extensive model training at test time.
  
\section{Quality improvement of generated images}
\label{sec:c3}
    Current generated images need general improvement. Generating high-quality and photorealistic images remains a challenge, and it often necessitates long and specialized prompts to achieve satisfactory results. Moreover, generated human faces may exhibit distortions or possess an oil-painting-like quality, indicating the need for further advancements in the image generation process.
    
    While larger model sizes and datasets can indeed improve performance, researchers have explored alternative approaches beyond simply scaling up these factors to enhance the existing framework in a more effective and efficient manner.
    
\subsection{Improve from the Text Encoder}    
    The research conducted by Imagen~\cite{NEURIPS2022_imagen} emphasizes the significant role of the text encoder in text-to-image generation models. According to their findings, enhancing the performance of the text encoder or increasing its size holds greater importance than optimizing the U-net component. In their study, Imagen~\cite{NEURIPS2022_imagen} explored various sizes and types of text encoders, and the results indicated that the T5-XXL text encoder~\cite{2020t5} outperformed smaller T5 text encoders and similar-sized CLIP text encoders. This improvement in performance was evident in the transition from SD 1.5 to SD 2.1, and also supported by experiments conducted on the GlueGen~\cite{qin2023gluegen}. These findings highlight the crucial influence of the text encoder in enhancing the overall performance and quality of text-to-image generation models.
    
    In the landscape of current multi-modal models, two prominent categories of text encoders are widely used: CLIP~\cite{pmlr-v139-clip} based text encoder and pure language pretrained text encoder like Bert~\cite{devlin2019bert}, Roberta~\cite{liu2019roberta}, t5~\cite{2020t5}. CLIP~\cite{pmlr-v139-clip} CLIP is specifically trained to maximize the mutual information between text and images, endowing its text encoder with a distinct advantage in text and image-related tasks. However, it appears that this alignment training is the only advantage of CLIP. Works like~\cite{ALBEF, dou2022meter, yang2022vision, yuksekgonul2023when}, all mentioned that the text encoder and the image encoder alone did not have any advantage compared with single modal models. In fact, straightforwardly speaking, Roberta~\cite{liu2019roberta} may serve as a more effective text encoder, providing richer semantic information than CLIP~\cite{pmlr-v139-clip}. An alternative approach was taken by eDiff-i~\cite{balaji2022eDiff-I}, which pursued a different path by employing not just one, but two text encoders: T5-XXL~\cite{2020t5} and CLIP~\cite{pmlr-v139-clip}. Remarkably, this hybrid text encoder configuration achieved the best results, showcasing the potential benefits of incorporating multiple text encoders for improved performance in multi-modal tasks.

    StructureDiffusion~\cite{feng2023structurediff} takes a different direction from the pursuit of a better text encoder. Their methodology centers on analyzing the generated text embedding, under the assumption that the self-attention mechanism can introduce perturbations to the text embedding. Specifically, they identify the occurrence of attribute mixing, where certain attributes, such as colors, may inadvertently be associated with incorrect objects in the sentence, leading to inaccurate image generation. To address this issue, StructureDiffusion analyzes the sentence structure and isolates object-attribute pairs. These pairs are then passed through the same text encoder, and the corresponding embeddings in the original sentence are replaced with the uncorrupted embeddings from the isolated pairs. For example, in the sentence ``a yellow apple and a green banana,'' the pairs ``yellow apple'' and ``green banana'' would be processed by the text encoder separately. The uncorrupted embedding from ``yellow apple'' would replace the original ``yellow'' attribute in the sentence, avoiding the mixing with ``green.'' The experiments conducted by StructureDiffusion demonstrate a reduction in attribute mixing cases, highlighting the effectiveness of their approach in enhancing the quality and accuracy of generated images by addressing potential perturbations in the text embeddings.
    
\subsection{Mixture of Experts}
    Mixture of experts (MOE)~\cite{MOE} is a technique that leverages the strengths of different models, and it has been adapted for use in diffusion models to optimize their performance. The generation process in a diffusion model involves thousands of steps, and these steps vary in nature. In the initial steps, the generation progresses from noise to form a rough image, while in the later steps, details are added to refine the image.
    
    ERNIE-ViLG~\cite{feng2023ernievilg} and eDiff-i~\cite{balaji2022eDiff-I} are two models that proposed to employ different models for different stages of the generation process. ERNIE-ViLG~\cite{feng2023ernievilg} uniformly divided the whole process in to several distinct stages,  with each stage being associated with a specific model. On the other hand, eDiff-i~\cite{balaji2022eDiff-I} calculated thresholds to separate the entire process into three stages. 

    The experiments conducted with ERNIE-ViLG~\cite{feng2023ernievilg} indicate that increasing the number of separated stages leads to improvements in performance. This highlights the effectiveness of employing a mixture of experts approach in diffusion models, allowing for better utilization of different models at various stages of the generation process, and ultimately enhancing the quality of the generated images.
    
\subsection{Instruction Tuning for Human Preference}
    The success of ChatGPT has sparked significant interest in instruction tuning~\cite{ouyang2022instructgpt, chung2022flan}. A methodology that naturally extends to image generation tasks due to diffusion model's Markov Process nature. In the context of image generation, the core concept of instruction tuning involves the following steps:
    \begin{itemize}
        \item Human-labeled data collection: The process begins with the collection of human-labeled data, forming a training set that reflects human preferences and quality assessments of generated images. This data serves as a crucial foundation for guiding the image generation process.
        \item Reward model training: A reward model is then trained using the collected human-labeled data. This model aims to capture and represent the preferences and evaluation criteria of humans regarding image quality. It serves as a reference for assessing the desirability of generated images.
        \item Reinforcement Learning (RL): Leveraging reinforcement learning techniques, the image generation process is guided by the reward model. The RL framework optimizes the overall preference and quality of the generated images by directing the generation process to maximize the rewards provided by the trained reward model.
    \end{itemize}

    By following these steps, instruction tuning can effectively enhance the quality of generated images, aligning the output with human preferences and quality standards, and thus providing a more refined and satisfying image generation experience.
    
    Indeed, several works such as~\cite{lee2023aligning,wu2023better,xu2023imagereward} have trained their own reward models to assess the human preference on the images. Additionally, the aesthetic score predictor model released by LAION can serve as a reward model, as demonstrated in~\cite{black2023training}. 

    Because the denoising process is already modeled as Markov Process, it is easy to fit it into RL framework. The denoising process is to predict $x_{t-1}$ conditioned on current $x_t$, time step $t$, and condition $c$, $p_\theta(X_{t-1}|X_t, t, c)$, the MarKov Decision Process can be modeled as:
    \begin{itemize}
        \item State: $X_t$,$t$,$c$
        \item Action: $X_{t-1}$ or the noise added.
        \item Reward: the output of the pre defined reward model
        \item Policy: $p_\theta(X_{t-1}|X_t, t, c)$
    \end{itemize}
    Given this formulation, any RL algorithm can be easily applied here. 
    
\subsection{Sampling Quality Improvement}
    Shifted Diffusion~\cite{zhou2022shifted} proposed another approach. By comprehensive study, they found out that, the distribution of image latent domain is only a very small subset of the whole embedding space. They recognized that in addition to noise removal, effectively guiding the samples towards this small subset of the image latent space is crucial for achieving high-quality image generation. To address this, they proposed the introduction of a shift term in the reverse process, giving rise to the method's distinctive name. After retraining the pipeline incorporating this modification, the experimental results demonstrated notable enhancements in the generated image quality.

\subsection{Self-Attention Guidance}
    SAG~\cite{hong2022SAG} also proposed their idea to improve the sample quality. Their methodology involves a thorough reexamination of the DDPM~\cite{NEURIPS2020_DDPM} process, leading to the introduction of a novel concept termed ``general diffusion guidance.'' Even in unconditioned image generation scenarios, they propose that the input white noise inherently carries valuable information that can guide the generation process, potentially resulting in the appearance of specific objects like an apple in the generated image.

    In their framework, they employ Gaussian blur on the on certain area of the prediction according to self-attention map to extract this condition, which is then used to guide the image generation process effectively. Their research reveals a strong correlation between the self-attention map and the detailed parts of the image. Consequently, this method enables the generation of images with heightened image details, contributing to a notable improvement in sample quality.

\subsection{Rewrite the Prompt}
    From the ``Civitai'' community, a very interesting conclusion is presented, if we write a very detailed prompt, the generate image quality can be improved dramatically. Detailed prompts here not only refer to a more detailed description of the objects, but also image quality related words like ``4K'', ``High resolution''. People from the community even systematically summarized how to write a good prompt. They said in a good prompt, for example to generate a image of a girl, you should have following components:
    \begin{itemize}
        \item the subject: such as ``a girl''
        \item detailed description: such as, the hair style, hair color, skin color, posture, clothing
        \item the environments or background: such as, ``in a coffee shop''
        \item the lighting: ``warm sunshine''
        \item the image quality related: ``4k'', ``master piece''
    \end{itemize}
    So, overall, the prompt became, ``a girl, with long straight hair, black hair, nike t-shirt, holding coffee in coffee shop, warm sunshine, 4k, master piece, detailed face, detailed hands''. 

\begin{figure}[t]
        \centering
        \includegraphics[width=0.45\textwidth]{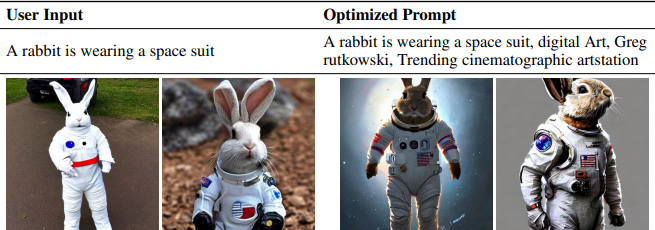}
        \caption{Prompt rewrite to get better image~\cite{hao2022Promptist}.}
        \vspace{-2mm}
        \label{fig:promptist}
    \end{figure}

    However, this is very hard for normal users to write such long and detailed prompt. Promptist~\cite{hao2022Promptist} proposed to finetune a language model, GPT2~\cite{radford2019gpt2} in this case, to rewrite the prompt and achieve better results, as illustrated in Figure \ref{fig:promptist}.

\section{Conclusion and Future Directions}
    {\zheng In this survey, we briefly introduce} the diffusion models, the current issues and the provided solutions. 
    {\zheng We organize the survey according to the three main challenges and the solutions on text-to-image generation, including} multi objects generation, rare case and unseen cases, and general improvement. However, it is important to note that image generation encompasses a broader scope beyond text-to-image generation. Some of the additional topics in image generation include: (1) Image editing: This task is straight forward. To modify the image, we can directly provide a prompt as an instruction~\cite{brooks2022instructpix2pix}, such as change the scene to nighttime. Or we can also interactive direct with the image like dragging as shown in DragGAN~\cite{pan2023_DragGAN,shi2023dragdiffusion} or Google's magic edit. Editing by generation model is more natural and more efficient compare with traditional method. (2) Inpainting: This task is given a image and masked a part of that image. The generation is to generate the masked part conditioned on the given prompts and the surrounding images. Adobe's Generative fill is the most famous work in this area, it is integrated in Photoshop.
    
    Future research in diffusion models should prioritize addressing the existing challenges and advancing the image generation process. While progress has been made, there are still gaps to be filled in order to achieve perfection in addressing the issues mentioned above. It is important to consider the inference time as a significant concern when comparing diffusion models to GANs. {\zheng Currently, few studies focus on tackling the challenges related to positional generation within the context of diffusion models.} Concept customization quality is still not good enough. Generally, to generate a perfect image, we still need to try a lot of times and the image quality can not be photo realistic. Additionally, ensuring the prevention of generating potential discriminating, toxic, or illegal content poses a considerable challenge. These areas necessitate the dedicated efforts of researchers to develop solutions and safeguards. 


\bibliographystyle{ACM-Reference-Format}
\bibliography{diffusion_ref}


\end{document}